\pdfoutput=1
\documentclass[11pt]{article}

\usepackage[final]{acl}

\makeatletter
\let\cas@beginabstract\abstract      % \abstract is the *begin* macro
\let\cas@endabstract  \endabstract   % \endabstract is the *end*  macro
\makeatother
\usepackage{arabtex}
\usepackage{utf8}

\makeatletter
\let\abstract   \cas@beginabstract   % restore begin
\let\endabstract\cas@endabstract     % restore end
\makeatother
\usepackage{times}
\usepackage{latexsym}
\usepackage{amssymb}
\usepackage{pifont}
\usepackage{multirow}
\usepackage{booktabs}
\usepackage{array}
\usepackage{graphicx}
\usepackage{xcolor}    % For color markers
\usepackage{tcolorbox}
\newcommand{\xmarkred}{\textcolor{red}{$\times$}}

\usepackage[T1]{fontenc}
\usepackage[utf8]{inputenc}

\usepackage{microtype}

\usepackage{inconsolata}

\usepackage{graphicx}

\title{From Guidelines to Practice: A New Paradigm for Arabic Language Model Evaluation}

\author{
\textbf{Serry Sibaee\textsuperscript{1*}} \quad
\textbf{Omer Nacar\textsuperscript{1}} \quad
\textbf{Adel Ammar\textsuperscript{1}} \quad
\textbf{Yasser Al-Habashi\textsuperscript{1}} \\
\textbf{Abdulrahman Al-Batati\textsuperscript{1}} \quad
\textbf{Wadii Boulila\textsuperscript{1}} \\
\textsuperscript{1}Prince Sultan University, Riyadh, Saudi Arabia \\
\texttt{\{sibaee, onajar, aammar, yalhabashi, aalbatati, wboulila\}@psu.edu.sa} \\
\textsuperscript{*}Corresponding author: \texttt{ssibaee@psu.edu.sa}
}

\setcode{utf8}
\begin{document}
\maketitle

\begin{abstract}
This paper addresses critical gaps in Arabic language model evaluation by establishing comprehensive theoretical guidelines and introducing a novel evaluation framework. We first analyze existing Arabic evaluation datasets, identifying significant issues in linguistic accuracy, cultural alignment, and methodological rigor. To address these limitations in LLMs, we present the Arabic Depth Mini Dataset (ADMD), a carefully curated collection of 490 challenging questions spanning ten major domains (42 sub-domains, see Figure \ref{fig:academic-categories}). Using ADMD, we evaluate five leading language models: GPT-4, Claude 3.5 Sonnet, Gemini Flash 1.5, CommandR 100B, and Qwen-Max. Our results reveal significant variations in model performance across different domains, with particular challenges in areas requiring deep cultural understanding and specialized knowledge. Claude 3.5 Sonnet demonstrated the highest overall accuracy at 30\%, showing relative strength in mathematical theory in Arabic, Arabic language, and islamic domains. This work provides both theoretical foundations and practical insights for improving Arabic language model evaluation, emphasizing the importance of cultural competence alongside technical capabilities.
\end{abstract}

\section{Introduction}
The evaluation of Arabic large language models (LLMs) presents unique challenges that extend beyond conventional metrics of linguistic accuracy. As these models become increasingly prevalent in various applications, the need for comprehensive and culturally aware evaluation frameworks has become critical. Recent developments in Arabic LLM evaluation have produced several datasets, including GPTArEval \cite{khondaker-etal-2023-gptaraeval}, Ghafa \cite{almazrouei-etal-2023-alghafa}, and ArabicMMLU from openAI \cite{openai2024mmmlu}, each attempting to address different aspects of model assessment. However, these efforts often fail to provide a comprehensive evaluation that includes both technical proficiency and cultural understanding.

\begin{figure}[t]
    \centering
    \includegraphics[width=0.5\textwidth]{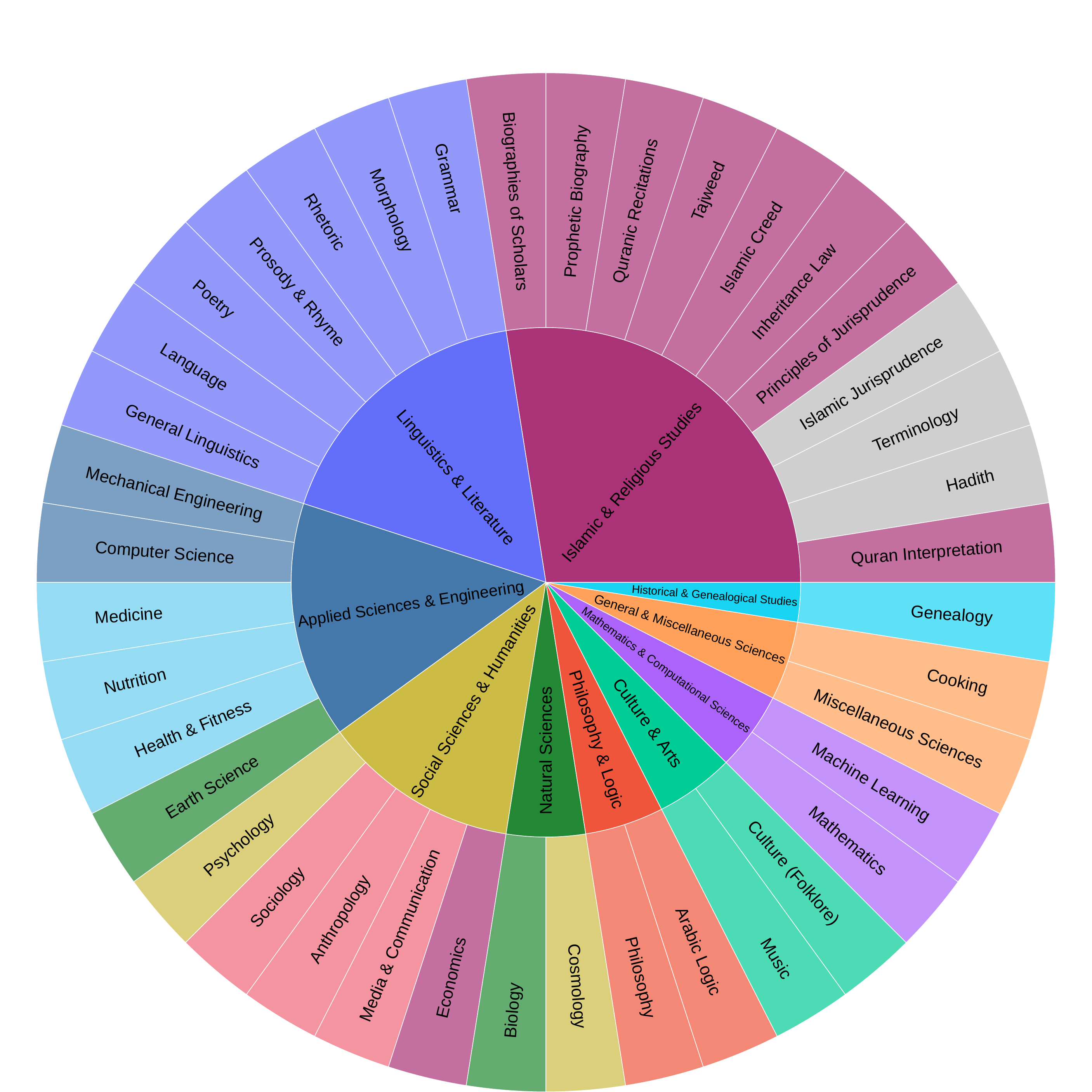}
    \caption{Representation of categories and subcategories of the proposed dataset.}
    \label{fig:academic-categories}
\end{figure}

Current evaluation approaches frequently rely on translated content \cite{romanou2024include} or simplified metrics that fail to capture the nuances of Arabic language and culture \cite{openai2024mmmlu}. This limitation is particularly evident in specialized domains such as Islamic studies, classical literature, and technical fields where cultural context and domain expertise are crucial. Furthermore, existing datasets often exhibit inconsistencies in linguistic standards and cultural representation, potentially resulting in misleading assessments of model capabilities.

Our work addresses these challenges through three main contributions. First, we establish theoretical guidelines for Arabic evaluation datasets that encompass linguistic standards, cultural alignment, and methodological requirements. Second, we conduct a detailed analysis of existing evaluation datasets, identifying common pitfalls and areas for improvement. Third, we introduce the Arabic Depth Mini Dataset (ADMD), a specialized evaluation tool designed to assess both technical and cultural competencies across diverse domains.

The ADMD represents a significant advancement in the evaluation of Arabic LLM, featuring carefully curated questions that demand deep understanding rather than surface-level pattern matching. By evaluating leading language models using this dataset, we provide insights into current model capabilities and limitations, particularly in handling complex Arabic queries that require cultural awareness and specialized knowledge.

This paper is organized as follows: Section 2 reviews related work in Arabic LLM evaluation, Section 3 presents our theoretical guidelines, Section 4 analyzes existing evaluation datasets, Section 5 introduces the ADMD and presents evaluation results, and Section 6 discusses limitations and future work directions. 

\begin{table*}[ht]
\centering
\resizebox{\textwidth}{!}{
\begin{tabular}{c c c c c}
\hline
\textbf{Dataset} & \textbf{Reviewed} & \textbf{Handwritten} & \textbf{Generated} & \textbf{Translated} \\  
\hline
GPTArEval~\cite{khondaker-etal-2023-gptaraeval} & \xmarkred & \checkmark & \xmarkred & \xmarkred \\  
Ghafa~\cite{almazrouei-etal-2023-alghafa} & \xmarkred & \checkmark & \xmarkred & \checkmark \\  
ArabicMMLU~\cite{koto-etal-2024-arabicmmlu} & \checkmark & \xmarkred & \xmarkred & \checkmark \\  
AraDICE~\cite{mousi-etal-2025-aradice} & \xmarkred & \checkmark & \xmarkred & \xmarkred \\  
ArSTEM~\cite{mustapha2024arastemnativearabicmultiple} & \checkmark & \xmarkred & \checkmark & \xmarkred \\  
Aya Expanse~\cite{dang2024ayaexpansecombiningresearch} & \xmarkred & \xmarkred & \checkmark & \checkmark \\  
AraTrust~\cite{alghamdi2024aratrustevaluationtrustworthinessllms} & \checkmark & \checkmark & \xmarkred & \xmarkred \\  
ILMAAM~\cite{nacar-etal-2025-towards} & \checkmark & \xmarkred & \checkmark & \xmarkred \\  
ArabicaQA~\cite{abdallah2024arabicaqacomprehensivedatasetarabic} & \checkmark & \checkmark & \xmarkred & \xmarkred \\  
Balsam~\cite{balsam2024} & \checkmark & \checkmark & \xmarkred & \xmarkred \\  
\textbf{ADMD (Ours)} & \textbf{\checkmark} & \textbf{\checkmark} & \textbf{\xmarkred} & \textbf{\xmarkred} \\  
\hline
\end{tabular}
}
\caption{Comparison of Arabic LLM Evaluation Datasets based on annotation type and content origin.}
\label{tab:arabic_llm_datasets}
\end{table*}

\section{Related Works}

The evaluation landscape for Arabic large language models (LLMs) has witnessed significant advancements through several benchmark initiatives, each with distinctive methodological approaches and inherent limitations \cite{eriksson2025trustaibenchmarksinterdisciplinary}. This section critically examines these evaluation frameworks while highlighting their methodological underpinnings and empirical contributions.

GPTArEval \cite{khondaker-etal-2023-gptaraeval} represents a pioneering effort in Arabic LLM assessment, with its integration of ORCA \cite{elmadany-etal-2023-orca} datasets and emphasis on natural language understanding and generation capabilities. The framework provides valuable insights into model performance but exhibits constraints in addressing the full spectrum of Arabic linguistic nuances. In parallel, Ghafa \cite{almazrouei-etal-2023-alghafa} employs a translation-based methodology supplemented by native speaker revisions, while ArabicMMLU \cite{koto-etal-2024-arabicmmlu} attempts to span diverse knowledge domains. Despite engaging ten native Arabic speakers in their validation processes, both datasets demonstrate substantial limitations in linguistic precision and comprehensive domain representation, with assessment complexity being constrained by the educational resources utilized in their development.

The cultural dimension of Arabic LLM evaluation has been addressed through specialized datasets such as AraDICE \cite{mousi-etal-2025-aradice}, which focuses specifically on dialectal variations and cultural contextual understanding, and ArSTEM \cite{mustapha2024arastemnativearabicmultiple}, which emphasizes scientific knowledge assessment within Arabic linguistic frameworks. These initiatives reflect an emerging recognition of the importance of culturally nuanced evaluation metrics that extend beyond mere linguistic accuracy.

Institutional contributions to Arabic LLM evaluation methodologies have come from teams developing models such as Jais \cite{sengupta2023jaisjaischatarabiccentricfoundation} and Allam \cite{bari2024allamlargelanguagemodels}, each implementing distinctive approaches to dataset curation and evaluation protocols. However, comprehensive assessment of closed-source models such as Allam and Fanar \cite{fanarteam2025fanararabiccentricmultimodalgenerative} remains challenging due to accessibility constraints. The Aya Expanse model \cite{dang2024ayaexpansecombiningresearch} distinguishes itself through methodological transparency regarding its utilization of translated and GPT-generated materials, establishing an important precedent for disclosure in evaluation dataset construction.

Critical methodological analysis by \cite{nacar-etal-2025-towards} has illuminated significant deficiencies in existing benchmarks, particularly in ArabicMMLU \cite{openai2024mmmlu}, encompassing linguistic inconsistencies, semantic imprecisions, and fundamental methodological flaws. In response to these identified shortcomings, AraTrust \cite{alghamdi2024aratrustevaluationtrustworthinessllms} was developed as a methodologically rigorous framework specifically designed to enhance reliability assessment for Arabic LLMs (as illustrated in Table \ref{tab:arabic_llm_datasets}). While \cite{abdallah2024arabicaqacomprehensivedatasetarabic} has contributed substantially to the field with an extensive Arabic question-answering dataset comprising over 80,000 entries authored by native speakers, its reliance on Wikipedia articles as source material raises concerns regarding authoritative credibility. In contrast, Balsam \cite{balsam2024} represents the highest quality dataset produced through collaboration between prominent academic and governmental institutions across the Middle East, though its utility is limited by the relatively small number of samples within each category and it is not publicly available.

Within the context of ongoing efforts to establish methodologically sound and linguistically accurate Arabic evaluation frameworks, our research makes two significant contributions: (1) a comprehensive theoretical analysis and empirical assessment of three critical evaluation datasets—Ghafa, ArabicMMLU, and INCLUDE—examining their methodological approaches, linguistic accuracy, and domain coverage; and (2) the introduction of the Arabic Depth Mini Dataset (ADMD), conceived as a foundational resource to address the current limitations in evaluating specialized domain knowledge within Arabic language contexts. The ADMD serves as an initial step toward developing a more extensive and rigorous Arabic question-answering dataset that can more effectively assess the depth of domain expertise in Arabic LLMs across disciplines.

\section{Theoretical Guidelines}

\begin{figure*}[h] % 'p' forces it to a separate page
    \centering
    \includegraphics[width=0.9\textwidth]{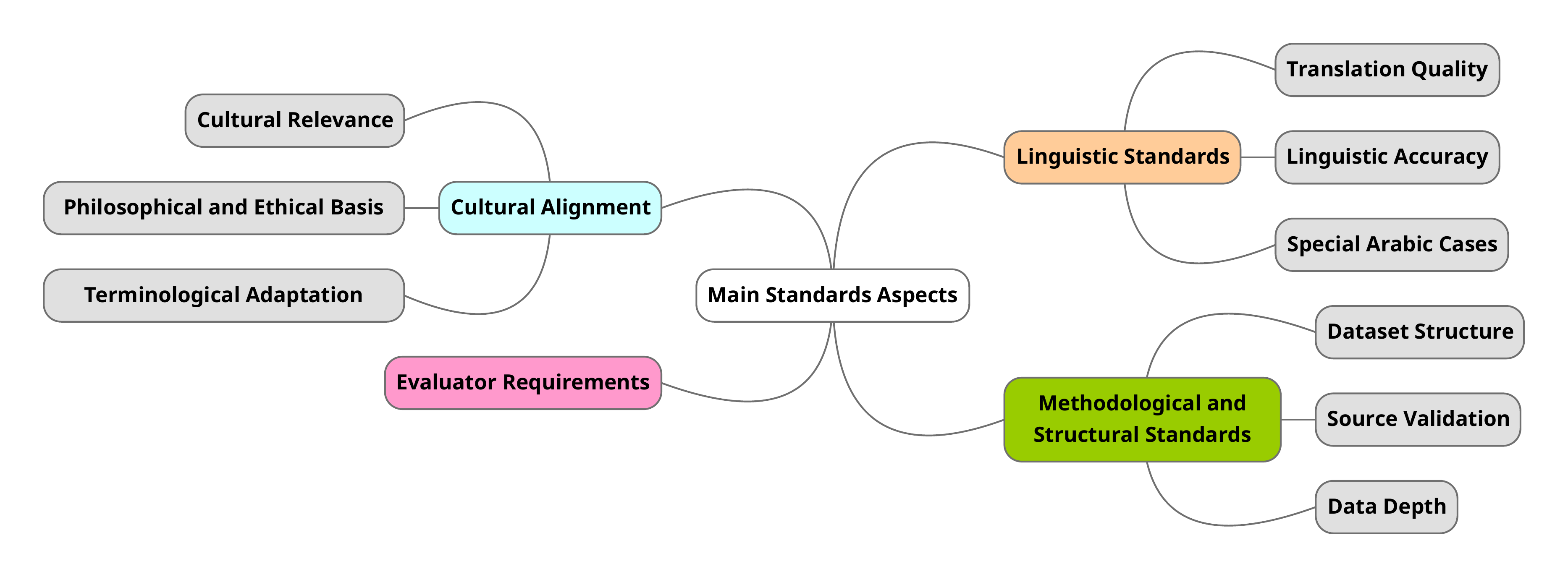} % Adjust width as needed
    \caption{Mindmap Representation of the theoretical standards}
    \label{fig:mindmap}
\end{figure*}

This section outlines the theoretical standards and instructions necessary for building an Arabic evaluation dataset, ensuring linguistic, cultural, and methodological soundness. The guidelines are categorized into  four areas: cultural, linguistics, methodology, and evaluation requirements  (Figure \ref{fig:mindmap}), and were inspired by the work of \cite{nacar-etal-2025-towards}.

\subsection{Linguistic Standards}
This section outlines the essential guidelines for ensuring high-quality and accurate translations, emphasizing linguistic precision, consistency, and contextual appropriateness in Arabic.
\begin{itemize}
    \item \textbf{Translation Quality:}
    \begin{itemize}
        \item Ensure that all terms are translated accurately; untranslated terms must be transliterated if necessary (and the non-Arabic word could be mentioned between brackets).
        \item Avoid literal translations by focusing on contextual adaptation, ensuring natural and consistent rendering.
        \item Review machine translations thoroughly and ensure alignment across multiple uses of the same term (e.g., consistency in letter choices for the answers like listing the Answers either in A,B,C or in Arabic \< أ, ب, ج>.
    \end{itemize}
    \item \textbf{Linguistic Accuracy:}
    \begin{itemize}
        \item Adhere strictly to Arabic grammar, morphology, syntax, and orthographic rules.
        \item Avoid weak linguistic structures even if grammatically correct.
        \item Ensure stylistic adequacy and use expressions that match the intended purpose and context.
    \end{itemize}
    \item \textbf{Special Cases:}
    \begin{itemize}
        \item Write poetry accurately, maintaining its structure and prosody.
        \item Write mathematical notations either in Arabic form or provide clear rules for using Latin symbols.
        \item Ensure consistent orthographic representation of dialects by adhering to a standard framework, for example, \cite{habash-etal-2018-unified} which gives standards to write Arabic dialects in a consistent way.
    \end{itemize}
\end{itemize}

\subsection{Cultural Alignment}
This subsection emphasizes aligning content with Arabic cultural contexts, adapting philosophical concepts, and using culturally appropriate terminology.
\begin{itemize}
    \item \textbf{Cultural Relevance:}
    \begin{itemize}
        \item Ensure questions, examples, and references align with the cultural, historical, and social contexts of the Arabic-speaking world.
        \item Avoid introducing examples or entities that are disconnected from Arab culture, such as irrelevant or Western-specific references.
    \end{itemize}
    \item \textbf{Philosophical and Ethical Basis:}
    \begin{itemize}
        \item Refrain from presenting Western philosophical or ethical concepts as universal truths without explanation or adaptation.
        \item Avoid using expressions or examples that conflict with the Arab cultural context or are confusing.
    \end{itemize}
    \item \textbf{Terminological Adaptation:}
    \begin{itemize}
        \item Replace Westernized terms with culturally and linguistically appropriate Arabic terms (in standard Arabic or in dialects).
        \item Provide Arabic equivalents or transliterations where necessary, maintaining cultural integrity.
    \end{itemize}
\end{itemize}

\subsection{Methodological and Structural Standards}
This subsection defines standards for organizing datasets, validating sources, and ensuring data depth and inclusivity.
\begin{itemize}
    \item \textbf{Dataset Structure:}
    \begin{itemize}
        \item Organize questions logically, ensuring they are placed in their relevant categories.
        \item Avoid redundancy or confusion by grouping related queries appropriately.
        \item Ensure the information is current and includes accurate dates.
    \end{itemize}
    \item \textbf{Source Validation:}
    \begin{itemize}
        \item Attribute knowledge and data to original Arabic primary sources, including books, studies, and statistical studies that are connected to Arabic societies.
        \item Avoid over-reliance on non-Arabic secondary references when constructing Arabic datasets.
        \item Writing Quranic texts with complete accuracy using the Uthmanic script. 
    \end{itemize}
    \item \textbf{Data Depth:}
    \begin{itemize}
        \item Ensure the dataset reflects depth and richness, avoiding straightforward, shallow, or overly simplistic questions and answers.
        \item Incorporate diverse perspectives within the Arabic-speaking world for inclusivity.
    \end{itemize}
\end{itemize}

% This subsection defines evaluator requirements, focusing on Arabic proficiency and subject matter expertise, and introduces a Python library for automated evaluation using the Claude Sonnet model.
% \begin{itemize}
%     \item Ensure that the evaluator or model being tested is not only understands Arabic, but is also proficient in the language's nuances and capable of grasping its cultural and historical contexts.
%     \item a solid understanding of the subject matter being evaluated.
% \end{itemize}
% A general mind map of the standards is displayed in Figure \ref{fig:mindmap}. and we Built a python library to automate the evaluation using Claude sonnet model\footnote{A side note the library was built after doing the full evaluation and analyzing the results -reason for choosing sonnet- the is on github: https://github.com/serrysibaee/EAED} to help evaluating evaluation datasets according to our theoretical guidelines.
\subsection{Evaluator Requirements}  
Evaluators must demonstrate proficiency in Arabic, encompassing both linguistic nuances and cultural contexts, complemented by solid subject matter expertise. Following comprehensive analysis, we developed a Python library that leverages the Claude Sonnet model to enhance the efficiency of the evaluation. This library, accessible via GitHub\footnote{\url{https://github.com/serrysibaee/EAED}}, automates dataset evaluation by applying theoretical guidelines. The model analyzes provided question/answer pairs and evaluates them according to specific criteria derived from established theoretical standards (for comprehensive details regarding the prompt implementation, please refer to Appendix A.3).

\section{Review of common Arabic Evaluation Datasets}

In this section, we review and assess three widely known Arabic evaluation datasets. A representative sample from each dataset was manually evaluated by one of the authors. The evaluation followed the theoretical framework proposed in the previous section and focused on four key criteria: \textbf{Language Rules}, \textbf{Scientific Writing}, \textbf{Cultural Values}, and \textbf{Information Correctness}. Each criterion was scored on a scale from 1 to 10.

\begin{itemize}
    \item \textbf{Language Rules:} This criterion refers to the proper use of Arabic grammar, syntax, and morphology. It includes the correctness of linguistic structures, agreement in gender and number, appropriate verb forms, and adherence to Standard Arabic norms.
    
    \item \textbf{Scientific Writing:} This evaluates the clarity, precision, and formality of the writing, particularly in scientific and technical contexts. It assesses whether the text follows the conventions of scientific discourse, including proper terminology usage, logical organization, and the avoidance of informal or ambiguous expressions.
    
    \item \textbf{Cultural Values:} This assesses the dataset’s sensitivity to cultural norms and values in Arabic-speaking communities. It considers inclusivity, the use of culturally appropriate examples, and the avoidance of content that may be considered offensive or misaligned with regional social norms.
    
    \item \textbf{Information Correctness:} This criterion examines the factual accuracy and consistency of the information provided in the dataset. It checks whether the content aligns with reliable knowledge sources and avoids misinformation or logical inconsistencies.
\end{itemize}

The datasets selected for evaluation are as follows:
\begin{enumerate}
    \item \textbf{Ghafa dataset} \cite{almazrouei-etal-2023-alghafa}, multiple-choice zero and few-shot evaluation benchmark across a wide range of tasks.
    
    \item \textbf{ArabicMMLU (OpenAI version)} \cite{openai2024mmmlu}, an Arabic adaptation of the MMLU benchmark, It covers a broad range of topics from 57 different categories, covering elementary-level knowledge up to advanced professional subjects like law, physics, history, and computer science..
    
    \item \textbf{Cohere "Include" dataset} \cite{romanou2024include}, an open-source multilingual dataset consists of 197,243 QA pairs from local exam sources to measure the capabilities of multilingual LLMs in a variety of regional contexts covering 44 languages including Arabic.
\end{enumerate}

Each dataset was assessed independently, and the results highlight both the progress made and the areas that still require improvement in Arabic evaluation benchmarks.

% \subsection{Al Ghafa Dataset}
% in this dataset \cite{almazrouei-etal-2023-alghafa} we sampled 100 examples from it and one native Arabic speaker reviewed it according to previous text. It scored [cite table]
% Lang Rules	Scien Writing	culture	Info correctness
% 4.53125	4.625	3.864583333	6.084210526

% and these are some examples from the evaluted samples 
\subsection{Al Ghafa Dataset}
From this dataset~\cite{almazrouei-etal-2023-alghafa}, we sampled 100 examples, which were reviewed by a native Arabic speaker according to the evaluation criteria outlined previously. The dataset received the evaluation scores shown in Table~\ref{tab:al_ghafa_scores}.

\begin{table}[h!]
\centering
\begin{tabular}{|l|c|}
\hline
\textbf{Criterion}            & \textbf{Score \textit{/10}} \\ \hline
Language Rules                & 4.5        \\ \hline
Scientific Writing            & 4.6          \\ \hline
Cultural Values               & 3.9         \\ \hline
Information Correctness       & 6.1         \\ \hline
\end{tabular}
\caption{Evaluation Scores for Al Ghafa Dataset}
\label{tab:al_ghafa_scores}
\end{table}
By considering that any sample that has an evaluation score lower than 5  is a 'wrong sample', we extracted: 50 wrong samples from language rules (Linguistic Accuracy), 42 from Scientific Writing, 60 from Cultural Values, and 26 from Information Correctness.
Below are examples of evaluated samples, along with their identified issues:

\begin{enumerate}
    \item \<صيام يوم الشك سنة>  
    \textbf{Translation:} Fasting on the day of doubt is a Sunnah.  
    \begin{itemize}
        \item \textbf{Issue:} The answer is inconsistent—its ruling depending on the disagreement.
    \end{itemize}

    \item \<(سَنَدْعُ الدْبَانِيَةَ (18))>
    \textbf{Translation:} Allah said, "So let him call his associates (17), We will call the guards of Hell (18)."  
    \begin{itemize}
        \item \textbf{Issue:} Incorrect transcription of the Quranic text, including errors in diacritics. The correct form is \<الزَّبانِيَة>.
    \end{itemize}

    \item \<الْـ 13 عَامْ بِيْتَرْ لِينْزْ>  
    \textbf{Translation:} Thirteen years old Peter Linz.  
    \begin{itemize}
        \item \textbf{Issue:} Grammatical error—the correct form is \<الْـ 13 عَامًا>.
    \end{itemize}

    \item \<كَمَا يَعْتَقِدُونَ فِيهِ العِصم ن هُوَ سَيِّدُ الرِّجَالِ؟>  
    \textbf{Translation:} As they believe in his infallibility, is he the master of men?  
    \begin{itemize}
        \item \textbf{Issue:} Spelling and typographical error. The correct form is \<الْعِصْمَةُ>.
    \end{itemize}
\end{enumerate}

\subsection{ArabicMMLU}
The Arabic MMLU Benchmark~\cite{openai2024mmmlu}, derived from the original English version~\cite{hendrycks2020measuring}, exists in two translations: one by GPT-3.5 Turbo and another by native Arabic translators. Despite its widespread adoption for Arabic LLM evaluation, the benchmark exhibits significant limitations in cultural adaptation and translation quality. Empirical analysis revealed three primary deficiencies:

 \textbf{(1): Linguistic Fidelity} following Arabic Grammar and translation quality,
 \textbf{(2): Cultural Alignment:} variant western focus with no Arabic alignment and
 \textbf{(3): Structural Integrity:} Suboptimal organization and insufficient Arabic source attribution. The evaluation scores are shown in Table \ref{tab:armmlu_resluts}.
\begin{table}[h!]
\centering
\begin{tabular}{|l|c|}
\hline
\textbf{Criterion}               & \textbf{Score \textit{/10}} \\ \hline
Language Rules                 & 6.5  \\ \hline
Scientific Writing                      & 5.5  \\ \hline
Culture Values                 & 3.4  \\ \hline
Information Correctness        & 6.5  \\ \hline
\end{tabular}
\caption{Evaluation Scores for ArabicMMLU Dataset.}
\label{tab:armmlu_resluts}
\end{table}

Below are three representative examples of identified issues:
\begin{enumerate}
    \item \<المضاعفات الفسيولوجية> \textbf{Translation:} Physiological complications
    \begin{itemize}
        \item \textbf{Issue:} did not translate the word \<الفسيولوجية> which has Arabic term \<الجسمورية، وظائف الأعضاء>.
    \end{itemize}

    \item \<المبادئ التوجيهية للجنة تكافؤ> \textbf{Translation}: Guidelines of the Equality Committee
    \begin{itemize}
        \item \textbf{Issue:} The reliance on Western laws and regulations without providing Arabic contextual alternatives. The solution is to train and use culturally aware models that understand the context and use the convenient words according to the Arabic culture.
    \end{itemize}

    \item No mention of studies or statistics of the Arabic society.

\end{enumerate}

\subsection{INCLUDE dataset}
INCLUDE~\cite{romanou2024include} is a multilingual benchmark evaluating knowledge and reasoning across 44 languages. The Arabic subset (551 MCQs) exhibited significant quality issues: (1) Poor Quality – 70\% contained severe spelling errors, and 80\% required major revisions in structure and content. (2) Incorrect Answers – Notably in Islamic studies, where precision is critical. (3) Misinformation – Some questions conveyed ambiguous or incorrect meanings, particularly in religious contexts. Table~\ref{tab:include} presents the dataset evaluation (excluding culture-related data\footnote{No culture data was in the dataset}).

\begin{table}[h!]
\centering
\begin{tabular}{|l|c|}
\hline
\textbf{Criterion}            & \textbf{Score \textit{/10}} \\ \hline
Language Rules                &   4.5      \\ \hline
Scientific Writing            &   3.5       \\ \hline
Cultural Values                &   -     \\ \hline
Information Correctness       &   7.0      \\ \hline
\end{tabular}
\caption{Evaluation Scores for INCLUDE Dataset.}
\label{tab:include}
\end{table}

Below are examples of evaluated samples along with their identified issues:

\begin{enumerate}
\item \textbf{Spelling Errors:}

\textbf{Original:} \<المنشؤة على تعّد> \textbf{Translation}: was constructed on.
    \begin{itemize}
        \item \textbf{Issue:} spelling mistake the correct is \<المنشأة على تعد>. 
    \end{itemize}

\item \textbf{Misleading Questions:}

\textbf{Original:} \<صوم رمضان سنة>
\textbf{Translation:} Fasting Ramadan is not mandatory.
\begin{itemize}
        \item\textbf{Issue:} Ramadan Fasting in Islam is mandatory.
    \end{itemize}

\end{enumerate}

% \section{MiniDataset}
% We developed a compact yet highly challenging Arabic dataset \footnote{uploaded to hugginface: } consisting of 490 questions, carefully curated from various books and resources. This dataset spans a wide range of advanced topics across ten major domains (more details in Appendix A):

% \begin{itemize}
%     \item \textbf{Applied Sciences \& Engineering}
%     \item \textbf{Natural Sciences}
%     \item \textbf{Social Sciences \& Humanities}
%     \item \textbf{Islamic \& Religious Studies}
%     \item \textbf{Linguistics \& Literature}
%     \item \textbf{Philosophy \& Logic}
%     \item \textbf{Culture \& Arts}
%     \item \textbf{Mathematics \& Computational Sciences}
%     \item \textbf{General \& Miscellaneous Sciences}
%     \item \textbf{Historical \& Genealogical Studies}
% \end{itemize}

% Unlike standard benchmarks, our evaluation is conducted through meticulous manual review rather than automated statistical analysis. We tested leading language models, including

% \begin{itemize}
%     \item GPT-4
%     \item Sonnet Claude 3.5 \footnote{claude-3-5-sonnet-20241022}
%     \item GEMINI FLASH 1.5
%     \item CommandR 100B \footnote{https://huggingface.co/CohereForAI/c4ai-command-r-plus}
%     \item Qwen-Max 2.5 \footnote{https://qwenlm.github.io/blog/qwen2.5-max/}
% \end{itemize}
%  to assess their ability to handle complex Arabic inquiries with precision and depth. main results are in Figure \ref{fig:models_results} and the main insights are presented in the next section.
\section{MiniDataset}
\begin{figure}[h!]
  \centering
  \includegraphics[width=0.5\textwidth]{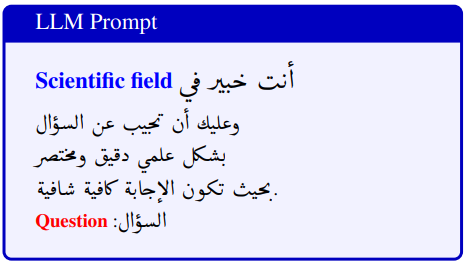} % Replace with your image filename
  \caption{The LLM prompt translates to: 'You are an expert in [Scientific field] and you need to answer the question scientifically and correctly. "Question"'. }
  \label{fig:LLM prompt}
\end{figure}

We developed (by our inner researchers in the lab: 3 Syrians, 1 Yemeni)  a compact yet highly challenging Arabic dataset\footnote{Data will be publicly available for a sample check appendix A.2} consisting of 490 carefully curated questions sourced from diverse books and references (more detail in Appendix A.2). The dataset spans ten major domains, covering general science, Islamic studies, Arabic language, and cultural topics (detailed in Appendix A). Unlike conventional benchmarks that rely on automated statistical analysis, our evaluation methodology is based on thorough manual review\footnote{After several experiments, we found that the most effective way to automate the evaluation is by using a judge LLM}. See Table \ref{words stats} for detailed statistics about the categories of the dataset, and Figure \ref{fig:LLM prompt} for the used prompt. We did not use LLM Judge in this paper because recent research shows \cite{wu2025bitterlessonlearned2000} that for non-English tasks, it is better to use manual evaluation.

% \begin{tcolorbox}[float, colback=blue!5!white, colframe=blue!75!black, title=LLM Prompt, label=box:llm_prompt]
% \textcolor{blue}{\textbf{[Scientific field]}} \<أنت خبير في> \\
% \small\<وعليك أن تجيب عن السؤال>
% \small\<بشكل علمي دقيق ومختصر>\\
% \small\<بحيث تكون الإجابة كافية شافية>.\\
%  \textcolor{red}{\textbf{Question}} :\<السؤال>
%  \label{LLM prompt}
% \end{tcolorbox}
% \noindent\begin{minipage}{\textwidth}
% \captionof{figure}{Prompt used for the LLM evaluation}\label{LLM prompt}
% \end{minipage}

\begin{figure*}[h]
    \centering
    \includegraphics[width=\textwidth]{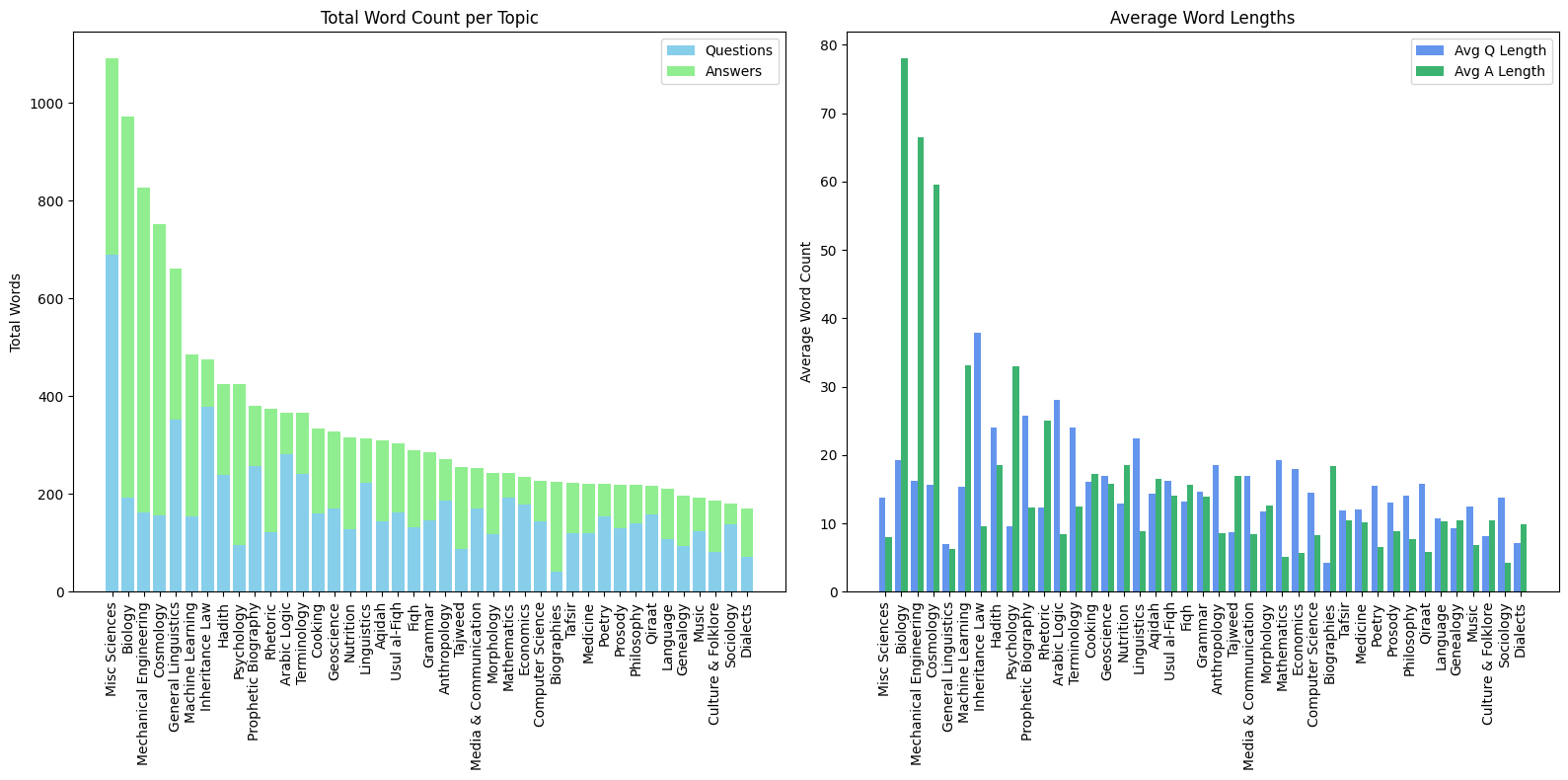}
    \caption{Visual summary of Q/A word counts}
    \label{fig:stats_graph}
\end{figure*}

\begin{figure}[h] % 'p' forces it to a separate page
    \centering
    \includegraphics[width=0.5\textwidth]{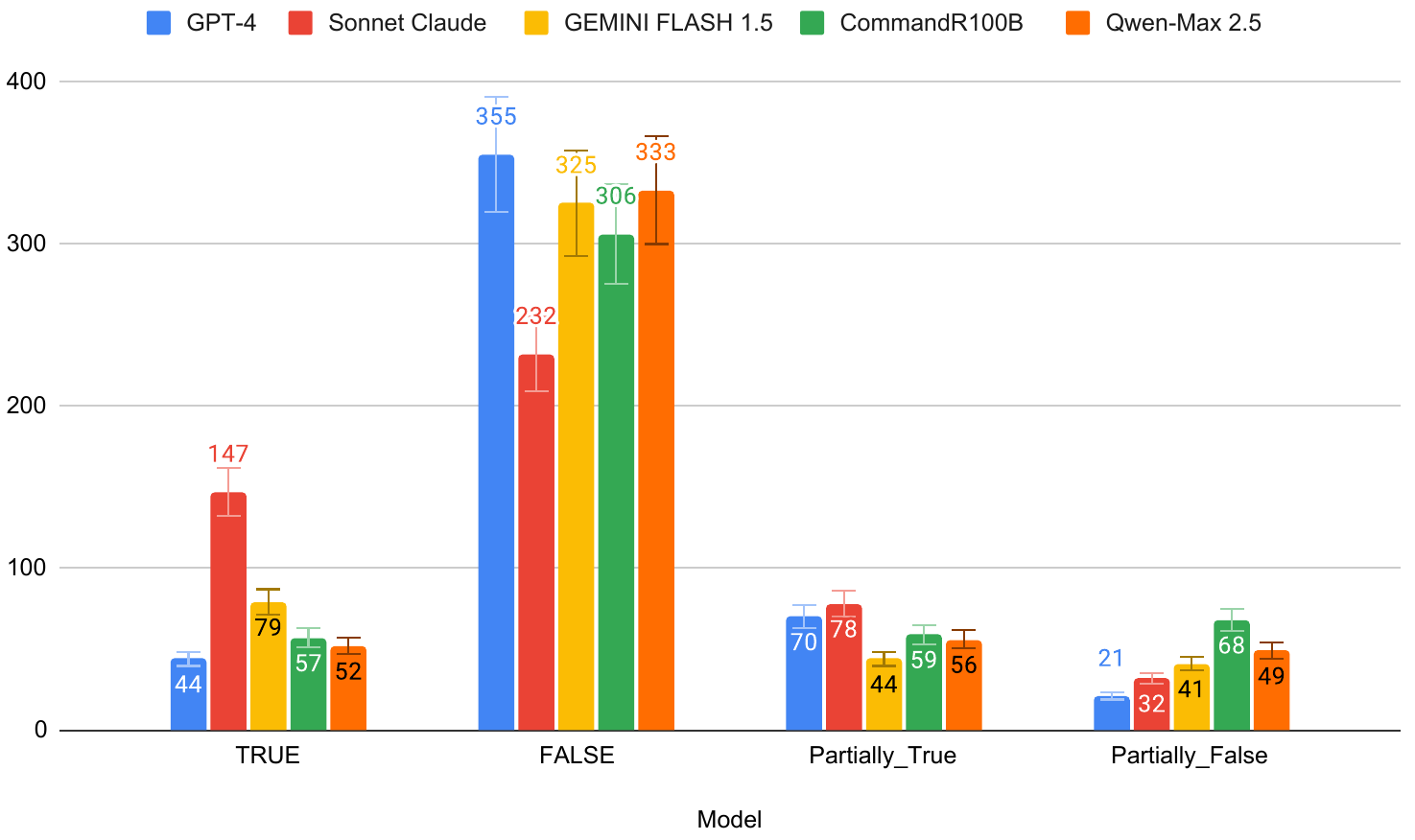} % Adjust width as needed
    \caption{Models' results. True means the model answered 100\% correctly and False means not-correct. Partially-True corresponds to an answer 60-80\% correct, whereas Partially-False corresponds to an answer only 20-30\% correct.}
    \label{fig:models_results}
\end{figure}

To assess the ability of language models to handle complex Arabic inquiries with precision and depth, we conducted extensive testing on leading models, including GPT-4, Sonnet Claude 3.5\footnote{claude-3-5-sonnet-20241022}, Gemini Flash 1.5, CommandR 100B\footnote{https://huggingface.co/CohereForAI/c4ai-command-r-plus}, and Qwen-Max 2.5\footnote{https://qwenlm.github.io/blog/qwen2.5-max/}. The primary results are presented in Figure \ref{fig:models_results}, with key insights discussed in the following section.

\subsection{Main insights}
The human evaluation results reveal significant performance differences among language models in handling complex Arabic questions\footnote{True means the model answered correctly and False is not-correct. Partially-True it
answered 60-80\% correct, Partially-False the answer is 20-30\% correct.}. \textbf{Claude 3.5 Sonnet} achieved the highest accuracy, correctly answering 147 questions (30\%), with notable strength in \textbf{Mathematics \& Computational Sciences (50\%)}, \textbf{Philosophy \& Logic (50\%)}, and \textbf{General \& Miscellaneous Sciences (51.67\%)}, as shown in Table~\ref{tab:stats_claude}. In \textbf{Natural Sciences}, it exhibited a balanced mix of \textbf{True (45\%)} and \textbf{Partially-True (45\%)} responses.

\textbf{GPT-4} had the weakest performance, with only 44 correct answers and the highest incorrect count (355) (Table~\ref{tab:gpt}), indicating difficulty in nuanced Arabic queries. \textbf{Gemini Flash 1.5} and \textbf{CommandR100B} showed moderate performance but high false rates (Table~\ref{tab:stats_gemini}, Table~\ref{tab:stats_commandR}). \textbf{Qwen-Max} had one of the lowest \textbf{True} counts (52) while being competitive in \textbf{Partially-True} responses (Table~\ref{tab:stats_qwen}), reflecting weaknesses in factual reasoning.

\textbf{Islamic \& Religious Studies} and \textbf{Linguistics \& Literature} had the highest \textbf{false rates}, with \textbf{Claude 3.5 Sonnet} performing relatively better (41.82\% False vs. over 80\% for other models). These results highlight the models' struggles with nuanced interpretation. Future improvements should focus on reducing \textbf{False} responses while refining \textbf{Partially-True} classifications to enhance factual accuracy.
\begin{table}[h!]
\centering
\small
\begin{tabular}{|l|c|c|c|c|}
\hline
\textbf{Model} & \textbf{T (\%)} & \textbf{F (\%)} & \textbf{PT (\%)} & \textbf{PF (\%)} \\ \hline
Sonnet 3.5 & \textbf{33.5} & \textbf{43.5} & 18.2 & 4.8 \\ \hline
Gemini & 22.1 & 56.2 & 12.0 & 9.7 \\ \hline
R+ & 15.0 & 54.0 & 15.6 & 15.4 \\ \hline
Qwen & 13.1 & 57.4 & 17.8 & 11.7 \\ \hline
GPT-4 & 11.8 & 67.3 & 17.3 & 3.5 \\ \hline
\end{tabular}
\caption{Model Performance Metrics (average for each model on the categories). T: True, F: False, PT: Partially-True, PF: Partially-False. True means the model answered 100\% correctly and False means not-correct. Partially-True corresponds to an answer 60-80\% correct, whereas Partially-False corresponds to an answer only 20-30\% correct.}
\label{tab:model_performance_admd}
\end{table}

\section{Limitations}
% \begin{itemize}
% \item \textbf{Evaluation Scalability}: Manual answer quality assessment is time-intensive, hindering scalability for larger datasets.
% \item \textbf{Limited Query Diversity}: The evaluation relied on a small number of questions per topic, potentially underrepresenting topic diversity.
% \item \textbf{Incomplete Topic Scope}: Key topics (e.g., Physics, Chemistry, Politics) and mathematical areas (e.g., Algebra, Topology) were excluded.
% \item \textbf{Minimal Field-Specific Testing}: Limited experimentation in fields like Medicine may skew performance insights.
% \item \textbf{Subjective Evaluation}: Topics such as Psychology and Cosmology involve subjectivity, complicating answer assessment.
% \item \textbf{Dataset Evaluation Overhead}: The dataset evaluation process was time-consuming and could be optimized.
% \item \textbf{Exclusion of Arabic Models}: Several notable Arabic models were not included, narrowing the scope of comparative analysis.
% \end{itemize}
The study faces several limitations, including the scalability challenge of manual evaluation and limited query diversity per topic. Key subjects such as Physics, Chemistry, and advanced mathematics were excluded, alongside minimal expertise in specialized fields like Medicine. Subjective topics (e.g., psychology, sociology) complicate assessment, and dataset evaluation remains time-intensive. Additionally, the exclusion of several Arabic models restricts the breadth of comparative analysis.

\section{Future Work}
% \begin{itemize}
%     \item \textbf{Dataset Expansion}: Future work will aim to expand the dataset to include a wider range of topics and an increased number of questions, providing a more comprehensive evaluation of model performance across various domains.
%     \item \textbf{Diverse Question Types}: Introducing different types of questions, such as Multiple Choice Questions (MCQs) and logic-based thinking questions, will help assess the models’ versatility in handling various question formats.
%     \item \textbf{Model Evaluation Expansion}: More models will be evaluated to provide a broader comparison, especially including state-of-the-art models like Jais, Allam, Fanar, Aya, and DeepSeek, as well as other high-performing Arabic models.
%     \item \textbf{Enhanced Prompting Strategies}: Future work will explore novel and optimized prompting techniques to improve the quality and accuracy of the models' responses, aiming to achieve better answers to complex questions.
% \end{itemize}
Future work will focus on expanding the dataset to cover more topics and question types, including MCQs and logic-based questions, to enhance evaluation comprehensiveness. Additional models, such as Jais \cite{sengupta2023jaisjaischatarabiccentricfoundation}, Allam \cite{bari2024allamlargelanguagemodels}, Fanar \cite{fanarteam2025fanararabiccentricmultimodalgenerative}, Aya \cite{dang2024ayaexpansecombiningresearch}, and DeepSeek \cite{liu2024deepseek}, will be assessed for broader comparison. Moreover, optimized prompting strategies will be explored to improve response accuracy and quality.
% \subsection{Future Work}
% \begin{itemize}
%     \item \textbf{Dataset Expansion}: Future work will aim to expand the dataset to include a wider range of topics and an increased number of questions, providing a more comprehensive evaluation of model performance across various domains.
%     \item \textbf{Diverse Question Types}: Introducing different types of questions, such as Multiple Choice Questions (MCQs) and logic-based thinking questions, will help assess the models’ versatility in handling various question formats.
%     \item \textbf{Model Evaluation Expansion}: More models will be evaluated to provide a broader comparison, especially including state-of-the-art models like Jais, Allam, Fanar, Aya, and DeepSeek, as well as other high-performing Arabic models.
%     \item \textbf{Enhanced Prompting Strategies}: Future work will explore novel and optimized prompting techniques to improve the quality and accuracy of the models' responses, aiming to achieve better answers to complex questions.
% \end{itemize}
\section{Conclusion}  
This paper proposed a comprehensive framework for evaluating Arabic language models, addressing linguistic, cultural, and methodological aspects. Our analysis identified limitations in existing evaluation datasets, including linguistic inaccuracies and cultural misalignment. To bridge these gaps, we introduced the Arabic Depth Mini Dataset (ADMD) with 490 questions across ten domains. Model evaluations using ADMD revealed varied performance, with Claude 3.5 Sonnet excelling in Mathematics \& Logic but all models struggling with culturally nuanced topics. These findings highlight the need for more refined evaluation methodologies to enhance Arabic NLP, ensuring both technical precision and cultural competence.

\section*{Acknowledgments}
The authors thank Prince Sultan University for their support.
\newpage
\bibliography{custom}
\appendix
\newpage
\newpage
\section{Appendix}
% \label{sec:appendix}
\subsection{Topics and References}

This dataset covers \textbf{42 topics} across various domains (each topic has 10 questions except general language and diversified science, which each have 50). The topics and their corresponding references are as follows (all of the used resources are publicly available books and websites):

\begin{itemize}
    \item \textbf{Applied Sciences \& Engineering:}  
    \begin{itemize}
        \item Mechanical Engineering  
        \begin{itemize}
            \item \textit{Asasiyat al-Ladoona} \footnote{\url{https://research-solution.com/}}  
            \item \textit{Al-Tibaa al-Thulathiyya al-Ab'ad} \footnote{\url{https://www.hindawi.org/books/16479648/}}  
        \end{itemize}
        \item Computer Science  
        \begin{itemize}
            \item \textit{Uloom al-Computer: Muqaddima Mukhtasara Jiddan} \footnote{\url{https://www.hindawi.org/books/13726364/}}  
        \end{itemize}
        \item Medicine  
        \begin{itemize}
            \item \textit{Mu'jam al-Mustalahat al-Tibbiyya (al-Qahira)}  
            \item \textit{Mawsu'at al-Tibb al-Nabawi li al-Asfahani}  
        \end{itemize}
        \item Nutrition (include Health \& Fitness)  
        \begin{itemize}
            \item \textit{Zad al-Ma'ad} \footnote{\url{https://shamela.ws/book/21713}}  
            \item \textit{Al-Sihha wa al-Taghziya – Dr. Iman Basheer Abukibda}  
            \item \textit{Kitab al-Tabikh} \footnote{\url{https://www.hindawi.org/books/93091736/}}  
        \end{itemize}
        \item Earth Science  
        \begin{itemize}
            \item \textit{Al-Mu'jam al-Geologi al-Musawwar} \footnote{\url{https://sgs.gov.sa/}}  
        \end{itemize}
    \end{itemize}

    \item \textbf{Natural Sciences:}  
    \begin{itemize}
        \item Biology  
        \begin{itemize}
            \item \textit{Ilm al-Ahyaa al-Takhalluqiyy: Muqaddima Qasira Jiddan} \footnote{\url{https://www.hindawi.org/books/42830286/}}  
            \item \textit{Suluk al-Hayawan} \footnote{\url{https://www.hindawi.org/books/86942829/}}  
        \end{itemize}
        \item Cosmology  
        \begin{itemize}
            \item \textit{Ilm al-Kawniyyat: Muqaddima Qasira Jiddan} \footnote{\url{https://www.hindawi.org/books/49725968/}}  
        \end{itemize}
    \end{itemize}

    \item \textbf{Social Sciences \& Humanities:}  
    \begin{itemize}
        \item Psychology  
        \begin{itemize}
            \item \textit{Mabadi' al-Tahlil al-Nafsi} \footnote{\url{https://www.hindawi.org/books/72426903/}}  
            \item \textit{Diraasat fi al-Takamul al-Nafsi} \footnote{\url{https://www.hindawi.org/books/93638206/}}  
        \end{itemize}
        \item Sociology  
        \begin{itemize}
            \item \textit{Mawsu'at Ilm al-Ijtima' – Gordan Marshal}  
            \item \textit{Ilm al-Ijtima' 'inda al-Arab} \footnote{\url{https://asjp.cerist.dz/en/article/70499}}  
        \end{itemize}
        \item Anthropology  
        \begin{itemize}
            \item \textit{Mawsu'at Ilm al-Insan – Charlotte Seymour}  
        \end{itemize}
        \item Media \& Communication  
        \begin{itemize}
            \item \textit{Maqalat 'Ilmiyya fi 'Ulum al-I'lam wa al-Ittisal} \footnote{\url{https://www.alukah.net/culture/0/84274/}}  
            \item \textit{Nazariyyat al-I'lam wa al-Ittisal} \footnote{\url{https://pedia.svuonline.org/}}  
        \end{itemize}
        \item Economics  
        \begin{itemize}
            \item \textit{Muhadarat fi Mabadi' al-Iqtisad} \footnote{\url{https://www.hindawi.org/books/91480861/}}  
            \item \textit{Duroos Mubassata fi al-Iqtisad} \footnote{\url{https://www.hindawi.org/books/60808479/}}  
        \end{itemize}
    \end{itemize}

    \item \textbf{Islamic \& Religious Studies:}  
    \begin{itemize}
        \item Quranic Exegesis (Tafsir): \textit{Tafsir Ibn Jarir al-Tabari} \footnote{\url{https://shamela.ws/book/43}}  
        \item Hadith: \textit{Tarh al-Tathreeb Sharh al-Taqreeb – al-Iraqi} \footnote{\url{https://shamela.ws/book/11036}}  
        \item Mustalah: \textit{Fath al-Mugheeth Sharh Alfiyyat al-Hadith – al-Subki} \footnote{\url{https://shamela.ws/book/5963}}  
        \item Fiqh: \textit{al-Mughni li Ibn Qudamah} \footnote{\url{https://shamela.ws/book/8463}}  
        \item Usul al-Fiqh: \textit{Sharh Mukhtasar al-Tahrir – Ibn al-Najjar} \footnote{\url{https://shamela.ws/book/12019}}  
        \item Fara'id: \textit{Al-As'ilah al-Mi'at al-Mukhtara fi al-Fara'id} \footnote{\url{https://almwareeth.com/}}  
        \item Aqeedah: \textit{Sharh al-Tahawiyyah – Ibn Abi al-Izz} \footnote{\url{https://shamela.ws/book/8352}}  
        \item Tajweed: \textit{al-Tamheed fi al-Tajweed – al-Jazari} \footnote{\url{https://shamela.ws/book/8194}}  
        \item Qira’at: \textit{al-Nashr fi al-Qira’at al-Ashr – Ibn al-Jazari} \footnote{\url{https://shamela.ws/book/22642}}  
        \item Seerah: \textit{Sirat Ibn Hisham} \footnote{\url{https://shamela.ws/book/23833}}  
        \item Tarajim al-Rijal: \textit{Nuzhat al-Fudala’}, \textit{al-Mukhtar al-Masoon} – Muhammad al-Sharif  
    \end{itemize}

    \item \textbf{Linguistics \& Literature:}  
    \begin{itemize}
        \item Nahw: \textit{Sharh Ibn Aqeel 'ala Alfiyyat Ibn Malik} \footnote{\url{https://shamela.ws/book/9904}}  
        \item Sarf: \textit{Sharh Tasreef al-Izzi – al-Taftazani}  
        \item Balagha: \textit{Sharh Talkhees al-Miftah – al-Taftazani}  
        \item Arood \& Qawafi: \textit{Meezan al-Dhahab fi Sina’at Shi’r al-Arab} \footnote{\url{https://www.hindawi.org/books/50259706/}}  
        \item Poetry: \textit{Shi’r al-Shu’ara’ al-Sitta – al-Shantamri} \footnote{\url{https://shamela.ws/book/5449}}, \textit{Muntaha al-Talab fi Shi’r al-Arab – Ibn Maymoon}  
        \item Arabic Language: \textit{Taj al-‘Aroos Sharh al-Qamoos} \footnote{\url{https://shamela.ws/book/7030}}, \textit{Lisan al-‘Arab – Ibn Manzur} \footnote{\url{https://shamela.ws/book/1687}}  
        \item General Linguistics: \textit{Diwan al-Lugha al-‘Arabiyya} \footnote{\url{http://diwanalarabia.com/}}  
        \item Arabic Linguistics: \textit{Kitab Ittila' ‘ala al-Nazariyyat al-Lisaniyya wa al-Dalaliyya}  
        \item Arabic Logic: \textit{Sharh Matn al-Shamsiyyah – al-Katibi}  
    \end{itemize}

    \item \textbf{Philosophy \& Logic:}  
    \begin{itemize}
        \item Philosophy: \textit{Mawsu’at al-Falsafah – al-Badawi}  
    \end{itemize}

    \item \textbf{Culture \& Arts:}  
    \begin{itemize}
        \item Music: \textit{Kitab al-Musiqa – al-Farabi}, \textit{Kitab al-Musiqa al-Sharqiyya} \footnote{\url{https://www.hindawi.org/books/46319638/3/}}  
        \item Folklore \& Cultural Studies: Interviews with native speakers from Yemen, Syria, Saudi Arabia, and Algeria.  
    \end{itemize}

    \item \textbf{Mathematics \& Computational Sciences:}  
    \begin{itemize}
        \item Mathematics: \textit{Mu'jam al-Riyadiyyat – Majma’ Dimashq} \footnote{\url{https://arabacademy-sy.org/ar/page17418/}}  
        \item Machine Learning: \textit{Mu'jam Mustalahat ‘Ilm al-Bayanat wa al-Ta’allum al-‘Amiq – ‘Alaa Tu’aymah} \footnote{\url{https://dlarabic.com/}}  
    \end{itemize}

    \item \textbf{General \& Miscellaneous Sciences:}  
    \begin{itemize}
        \item General Sciences: \textit{Hindawi Science Books} \footnote{\url{https://www.hindawi.org/}}  
        \item Cooking: \textit{Kitab al-Tabikh} \footnote{\url{https://www.hindawi.org/books/93091736/}}  
    \end{itemize}

    \item \textbf{Historical \& Genealogical Studies:}  
    \begin{itemize}
        \item Genealogy (Ansab): \textit{Kitab al-Ansab – al-Sam’ani} \footnote{\url{https://shamela.ws/book/1656}}  
    \end{itemize}

    \item \textbf{Language Extensions:}
    \begin{itemize}
        \item General Arabic Language (50 questions): \textit{Diwan al-Lugha al-‘Arabiyya} \footnote{\url{http://diwanalarabia.com/}}  
        \item Diversified Sciences (50 questions): \textit{Hindawi Science Collection} \footnote{\url{https://www.hindawi.org/}}  
        \item Dialects (Lahajat): \textit{Mo3jam} \footnote{\url{https://ar.mo3jam.com/dialect/}}, \textit{‘Amiyah} \footnote{\url{https://3amyah.com/}}  
    \end{itemize}
\end{itemize}

This structured categorization ensures a well-organized representation of the dataset's diverse topics, making it suitable for evaluating Arabic LLMs across multiple domains.  

\subsection{Examples from the ADMD}

In this section, we present examples from each topic in the ADMD dataset. Due to the length of these examples and technical issues related to handling long Arabic texts in the ACL format, we have opted to provide the examples in a more accessible format via a Google Sheet. This allows for easier reading and also includes their English translations.

You can access the examples and their translations through the following link:

\url{https://docs.google.com/spreadsheets/d/1Nl9ZDzNK29yJPpFepx453Lhbwf6HAPSnc_K5sGIfZ7U/edit?usp=sharing}

\subsection{Prompt for Evaluating in EAED}

% ------------------------------------------------------
% Linguistic Standard Evaluation
% ------------------------------------------------------
This prompt evaluates the linguistic quality of Arabic texts—including grammar, style, and exceptions like poetry or dialects. It aims to ensure objective scoring based on strict language norms.

\begin{tcolorbox}[float, colback=blue!5!white, colframe=blue!75!black, 
  title=LLM Prompt, label=box:linguistic_eval, boxrule=0.5pt, 
  fonttitle=\scriptsize\bfseries, sharp corners=south]
\scriptsize
\textcolor{blue}{\textbf{Scientific field}} <You are an expert Arabic linguist and evaluator> \\
<You must evaluate the given Arabic text based on grammar, syntax, morphology, and stylistic clarity> \\
<Ensure scoring includes special cases such as poetry, dialect, and mathematical notation> \\
<Provide a final numeric score only, from 1 to 10>. \\
\textcolor{red}{\textbf{Question}} : <Evaluate the following Arabic text and return only a numeric score.> \\
\end{tcolorbox}

% ------------------------------------------------------
% Translation Evaluation
% ------------------------------------------------------
Assesses the quality of Arabic translations, focusing on grammar, meaning accuracy, and cultural alignment. It follows professional standards for rating fidelity and fluency.

\begin{tcolorbox}[float, colback=blue!5!white, colframe=blue!75!black, 
  title=LLM Prompt, label=box:translation_eval, boxrule=0.5pt, 
  fonttitle=\scriptsize\bfseries, sharp corners=south]
\scriptsize
\textcolor{blue}{\textbf{Scientific field}} <You are a highly skilled Arabic translator> \\
<Your task is to assess a translation's quality, consistency, and grammatical integrity> \\
<Evaluate special elements like poetic structure and notation formatting> \\
<Give a score from 1 to 10, printed alone>. \\
\textcolor{red}{\textbf{Question}} : <Assess the quality of this Arabic translation and return only a number>. \\
\end{tcolorbox}

% ------------------------------------------------------
% Cultural Evaluation
% ------------------------------------------------------
The following prompt assesses how well the text aligns with Arabic cultural and ethical standards, including terminology usage and relevance to the societal context.

\begin{tcolorbox}[float, colback=blue!5!white, colframe=blue!75!black, 
  title=LLM Prompt, label=box:cultural_eval, boxrule=0.5pt, 
  fonttitle=\scriptsize\bfseries, sharp corners=south]
\scriptsize
\textcolor{blue}{\textbf{Scientific field}} <You are an expert in Arabic language and culture> \\
<You must assess the cultural relevance and sensitivity of the given text> \\
<Evaluate alignment with Arab societal, ethical, and terminological standards> \\
<Return only a number from 1 to 10>. \\
\textcolor{red}{\textbf{Question}} : <Evaluate this text's cultural alignment and print a numeric score>. \\
\end{tcolorbox}

% ------------------------------------------------------
% Methodology Evaluation
% ------------------------------------------------------
The following prompt focuses on structure, credibility, and richness of Arabic data or texts. It ensures sources are validated and the methodology is coherent and rooted in authentic Arabic references.

\begin{tcolorbox}[colback=blue!5!white, colframe=blue!75!black, 
  title=LLM Prompt, label=box:methodology_eval, boxrule=0.5pt, 
  fonttitle=\scriptsize\bfseries, sharp corners=south]
\scriptsize
\textcolor{blue}{\textbf{Scientific field}} <You are an expert in Arabic data structuring and methodology> \\
<Evaluate the dataset or text for organization, source validation, and informational richness> \\
<Verify alignment with Arabic primary sources and proper use of terminology> \\
<Respond with only a numeric score between 1 and 10>. \\
\textcolor{red}{\textbf{Question}} : <Assess the methodological quality of the following text or dataset and return only a number>. \\
\end{tcolorbox}

% \begin{figure*}[h] % 'p' forces it to a separate page
%     \centering
%     \includegraphics[width=0.9\textwidth]{arabic standards.pdf} % Adjust width as needed
%     \caption{Mindmap Representation}
%     \label{fig:mindmap}
% \end{figure*}

% \begin{figure*}[h] % 'p' forces it to a separate page
%     \centering
%     \includegraphics[width=0.9\textwidth]{models_results.pdf} % Adjust width as needed
%     \caption{Models Results: True means the model answered correctly and False is not-correct. Partially-True it answered 60-80\% correctly, Partially-False the answer is 20-30\% correct.}
%     \label{fig:models_results}
% \end{figure*}

\subsection{Detailed Tables}
\subsubsection{Details analysis about ADMD dataset QnAs}
This table is giving a comprehensive analysis about each category of the dataset with it number of rows, number of words in each Question and Answer, avrage words in Question and answer. 
\begin{table*}[h]
\centering
\footnotesize
\begin{tabular}{|l|r|r|r|r|r|r|}
\hline
\textbf{Subject} & \textbf{Rows} & \textbf{Q Words} & \textbf{A Words} & \textbf{Avg Q} & \textbf{Avg A} & \textbf{Total} \\
\hline
Machine Learning         & 10 & 154 & 331 & 15.40 & 33.10 & 485 \\
Tafsir (Exegesis)        & 10 & 119 & 105 & 11.90 & 10.50 & 224 \\
Grammar (Nahw)           & 10 & 146 & 139 & 14.60 & 13.90 & 285 \\
Psychology               & 10 & 96  & 330 & 9.60  & 33.00 & 426 \\
Nutrition                & 10 & 129 & 186 & 12.90 & 18.60 & 315 \\
Dialects                 & 10 & 71  & 99  & 7.10  & 9.90  & 170 \\
Mechanical Engineering   & 10 & 163 & 665 & 16.30 & 66.50 & 828 \\
Biology                  & 10 & 192 & 780 & 19.20 & 78.00 & 972 \\
Cosmology                & 10 & 156 & 596 & 15.60 & 59.60 & 752 \\
Biography                & 10 & 42  & 184 & 4.20  & 18.40 & 226 \\
Morphology (Sarf)        & 10 & 118 & 126 & 11.80 & 12.60 & 244 \\
Tajweed                  & 10 & 87  & 169 & 8.70  & 16.90 & 256 \\
Rhetoric (Balagha)       & 10 & 123 & 251 & 12.30 & 25.10 & 374 \\
Prosody                  & 10 & 131 & 88  & 13.10 & 8.80  & 219 \\
Jurisprudence (Fiqh)     & 10 & 132 & 157 & 13.20 & 15.70 & 289 \\
Inheritance (Fara'id)    & 10 & 379 & 96  & 37.90 & 9.60  & 475 \\
Usul al-Fiqh             & 10 & 163 & 140 & 16.30 & 14.00 & 303 \\
Creed (Aqidah)           & 10 & 144 & 165 & 14.40 & 16.50 & 309 \\
Hadith Terminology       & 10 & 241 & 125 & 24.10 & 12.50 & 366 \\
Hadith                   & 10 & 240 & 186 & 24.00 & 18.60 & 426 \\
Prophetic Biography      & 10 & 257 & 124 & 25.70 & 12.40 & 381 \\
Mathematics              & 10 & 192 & 51  & 19.20 & 5.10  & 243 \\
Linguistics              & 10 & 224 & 89  & 22.40 & 8.90  & 313 \\
Logic (Arabic)           & 10 & 281 & 85  & 28.10 & 8.50  & 366 \\
Genealogy                & 10 & 93  & 104 & 9.30  & 10.40 & 197 \\
Poetry                   & 10 & 155 & 66  & 15.50 & 6.60  & 221 \\
Qur’anic Readings        & 10 & 158 & 59  & 15.80 & 5.90  & 217 \\
Music                    & 10 & 125 & 68  & 12.50 & 6.80  & 193 \\
Folklore \& Culture       & 10 & 82  & 104 & 8.20  & 10.40 & 186 \\
Philosophy               & 10 & 141 & 77  & 14.10 & 7.70  & 218 \\
Geoscience               & 10 & 170 & 158 & 17.00 & 15.80 & 328 \\
Language (General)       & 10 & 107 & 103 & 10.70 & 10.30 & 210 \\
Economics                & 10 & 179 & 57  & 17.90 & 5.70  & 236 \\
Sociology                & 10 & 138 & 43  & 13.80 & 4.30  & 181 \\
Medicine                 & 10 & 121 & 101 & 12.10 & 10.10 & 222 \\
Computer Science         & 10 & 145 & 83  & 14.50 & 8.30  & 228 \\
Anthropology             & 10 & 186 & 86  & 18.60 & 8.60  & 272 \\
Media \& Communication    & 10 & 170 & 84  & 17.00 & 8.40  & 254 \\
General Linguistics      & 50 & 352 & 310 & 7.04  & 6.20  & 662 \\
General Sciences         & 50 & 689 & 402 & 13.78 & 8.04  & 1091 \\
Cooking                  & 10 & 161 & 173 & 16.10 & 17.30 & 334 \\
\hline
\end{tabular}
\caption{Word statistics across subjects (first 10 rows per sheet, or 50 for long sheets).}
\label{words stats}
\end{table*}

\newpage
\subsection{Detailed tables for LLM Evaluation}
These tables show detailed numbers for each model evaluation on the ADMD dataset. 

\begin{table*}[h]
\centering
\begin{tabular}{|l|c|c|c|c|}
\hline
\textbf{Field of Study} & \textbf{True (\%)} & \textbf{False (\%)} & \textbf{Partially-True (\%)} & \textbf{Partially-False (\%)} \\
\hline
Applied Sciences \& Engineering & 22.00 & 42.00 & 28.00 & 8.00 \\
\hline
Natural Sciences & 20.00 & 35.00 & 45.00 & 0.00 \\
\hline
Social Sciences \& Humanities & 12.00 & 56.00 & 26.00 & 6.00 \\
\hline
Islamic \& Religious Studies & 0.91 & 80.91 & 10.00 & 8.18 \\
\hline
Linguistics \& Literature & 1.82 & 94.55 & 2.73 & 0.91 \\
\hline
Philosophy \& Logic & 10.00 & 80.00 & 10.00 & 0.00 \\
\hline
Culture \& Arts & 10.00 & 75.00 & 10.00 & 5.00 \\
\hline
Mathematics \& Computational Sciences & 25.00 & 45.00 & 25.00 & 5.00 \\
\hline
General \& Miscellaneous Sciences & 16.67 & 65.00 & 16.67 & 1.67 \\
\hline
Historical \& Genealogical Studies & 0.00 & 100.00 & 0.00 & 0.00 \\
\hline
\end{tabular}
\caption{Statistics for GPT-4 answers for the different categories.}
\label{tab:gpt}
\end{table*}

\begin{table*}[h]
\centering
\begin{tabular}{|l|c|c|c|c|}
\hline
\textbf{Field of Study} & \textbf{True (\%)} & \textbf{False (\%)} & \textbf{Partially-True (\%)} & \textbf{Partially-False (\%)} \\
\hline
Applied Sciences \& Engineering & 20.00 & 42.00 & 18.00 & 20.00 \\
\hline
Natural Sciences & 20.00 & 15.00 & 40.00 & 25.00 \\
\hline
Social Sciences \& Humanities & 18.00 & 42.00 & 24.00 & 16.00 \\
\hline
Islamic \& Religious Studies & 4.55 & 80.00 & 5.45 & 10.00 \\
\hline
Linguistics \& Literature & 1.82 & 90.00 & 3.64 & 4.55 \\
\hline
Philosophy \& Logic & 15.00 & 70.00 & 5.00 & 10.00 \\
\hline
Culture \& Arts & 5.00 & 85.00 & 10.00 & 0.00 \\
\hline
Mathematics \& Computational Sciences & 20.00 & 30.00 & 35.00 & 15.00 \\
\hline
General \& Miscellaneous Sciences & 26.67 & 50.00 & 16.67 & 6.67 \\
\hline
Historical \& Genealogical Studies & 0.00 & 70.00 & 20.00 & 10.00 \\
\hline
\end{tabular}
\caption{Statistics for Qwen-Max.}
\label{tab:stats_qwen}
\end{table*}

\begin{table*}[h]
\centering
\begin{tabular}{|l|c|c|c|c|}
\hline
\textbf{Field of Study} & \textbf{True (\%)} & \textbf{False (\%)} & \textbf{Partially-True (\%)} & \textbf{Partially-False (\%)} \\
\hline
Applied Sciences \& Engineering & 30.00 & 52.00 & 6.00 & 12.00 \\
\hline
Natural Sciences & 30.00 & 15.00 & 50.00 & 5.00 \\
\hline
Social Sciences \& Humanities & 18.00 & 46.00 & 20.00 & 16.00 \\
\hline
Islamic \& Religious Studies & 3.64 & 69.09 & 10.00 & 17.27 \\
\hline
Linguistics \& Literature & 4.55 & 82.73 & 3.64 & 9.09 \\
\hline
Philosophy \& Logic & 10.00 & 45.00 & 15.00 & 30.00 \\
\hline
Culture \& Arts & 15.00 & 70.00 & 5.00 & 10.00 \\
\hline
Mathematics \& Computational Sciences & 25.00 & 30.00 & 25.00 & 20.00 \\
\hline
General \& Miscellaneous Sciences & 13.33 & 60.00 & 11.67 & 15.00 \\
\hline
Historical \& Genealogical Studies & 0.00 & 70.00 & 10.00 & 20.00 \\
\hline
\end{tabular}
\caption{Statistics for commandR\_100B.}
\label{tab:stats_commandR}
\end{table*}

\begin{table*}[h]
\centering
\begin{tabular}{|l|c|c|c|c|}
\hline
\textbf{Field of Study} & \textbf{True (\%)} & \textbf{False (\%)} & \textbf{Partially-True (\%)} & \textbf{Partially-False (\%)} \\
\hline
Applied Sciences \& Engineering & 24.00 & 46.00 & 24.00 & 6.00 \\
\hline
Natural Sciences & 40.00 & 15.00 & 20.00 & 25.00 \\
\hline
Social Sciences \& Humanities & 38.00 & 32.00 & 14.00 & 16.00 \\
\hline
Islamic \& Religious Studies & 0.00 & 88.18 & 5.45 & 6.36 \\
\hline
Linguistics \& Literature & 2.75 & 84.40 & 4.59 & 8.26 \\
\hline
Philosophy \& Logic & 15.00 & 60.00 & 5.00 & 20.00 \\
\hline
Culture \& Arts & 10.00 & 70.00 & 10.00 & 10.00 \\
\hline
Mathematics \& Computational Sciences & 45.00 & 30.00 & 25.00 & 0.00 \\
\hline
General \& Miscellaneous Sciences & 36.67 & 56.67 & 1.67 & 5.00 \\
\hline
Historical \& Genealogical Studies & 10.00 & 80.00 & 10.00 & 0.00 \\
\hline
\end{tabular}
\caption{Statistics for Gemini-1.5-flash.}
\label{tab:stats_gemini}
\end{table*}

\begin{table*}[h]
\centering
\begin{tabular}{|l|c|c|c|c|}
\hline
\textbf{Field of Study} & \textbf{True (\%)} & \textbf{False (\%)} & \textbf{Partially-True (\%)} & \textbf{Partially-False (\%)} \\
\hline
Applied Sciences \& Engineering & 42.00 & 28.00 & 24.00 & 6.00 \\
\hline
Natural Sciences & 45.00 & 5.00 & 45.00 & 5.00 \\
\hline
Social Sciences \& Humanities & 38.00 & 38.00 & 20.00 & 4.00 \\
\hline
Islamic \& Religious Studies & 30.00 & 41.82 & 16.36 & 11.82 \\
\hline
Linguistics \& Literature & 12.84 & 66.97 & 13.76 & 6.42 \\
\hline
Philosophy \& Logic & 50.00 & 50.00 & 0.00 & 0.00 \\
\hline
Culture \& Arts & 15.00 & 65.00 & 15.00 & 5.00 \\
\hline
Mathematics \& Computational Sciences & 50.00 & 20.00 & 20.00 & 10.00 \\
\hline
General \& Miscellaneous Sciences & 51.67 & 40.00 & 8.33 & 0.00 \\
\hline
Historical \& Genealogical Studies & 0.00 & 80.00 & 20.00 & 0.00 \\
\hline
\end{tabular}
\caption{Statistics for Claude-3-5-sonnet.}
\label{tab:stats_claude}
\end{table*}

\end{document}